\documentclass{nature}

\usepackage[ruled,vlined]{algorithm2e}
\usepackage{amssymb}
\usepackage{xcolor}
\usepackage[utf8]{inputenc}

\title{Active learning of deep surrogates for PDEs: Application to metasurface design}
\author{Raphaël Pestourie$^{1,\ast}$, Youssef Mroueh$^{2,3}$, Thanh V. Nguyen$^{4}$, Payel Das$^{2,3,\ast}$, Steven G. Johnson$^1$
}

\date{\today}

\usepackage{cite}
\newcommand{\citeasnoun}[1]{Ref.~\citenum{#1}}
\newcommand{\secref}[1]{Sec.~\ref{sec:#1}}

\newcommand{\eqref}[1]{Eq.~(\ref{eq:#1})}
\newcommand{\Eqref}[1]{Equation~(\ref{eq:#1})}

\usepackage{graphicx}
\usepackage{url}
\begin{document}

\maketitle

\noindent \normalsize{$^{1}$ MIT, 77 Massachusetts Ave, Cambridge, MA 02139, USA }\\
\normalsize{$^{2}$ IBM Research AI, IBM Thomas J Watson Research Center,  Yorktown Heights, NY 10598, USA}\\
\normalsize{$^{3}$ MIT-IBM Watson AI Lab, Cambridge, MA 02139, USA}\\
\normalsize{$^{4}$ Iowa State University, Ames, IA 50011, USA }\\
\normalsize{$^\ast$Correspondence to: rpestour@mit.edu; daspa@us.ibm.com.}

\begin{abstract}

    Surrogate models for partial-differential equations are widely used in the design of metamaterials to rapidly evaluate the behavior of composable components. However, the training cost of accurate surrogates by machine learning can rapidly increase with the number of variables.  For photonic-device models, we find that this training becomes especially challenging as design regions grow larger than the optical wavelength.  We present an active learning algorithm that reduces the number of training points by more than an order of magnitude for a neural-network surrogate model of optical-surface components compared to random samples.  Results show that the surrogate evaluation is over two orders of magnitude faster than a direct solve, and we demonstrate how this can be exploited to accelerate large-scale engineering optimization.
\end{abstract}

\section{Introduction}

\begin{figure}[t!]
\centering
\includegraphics[width=\textwidth]{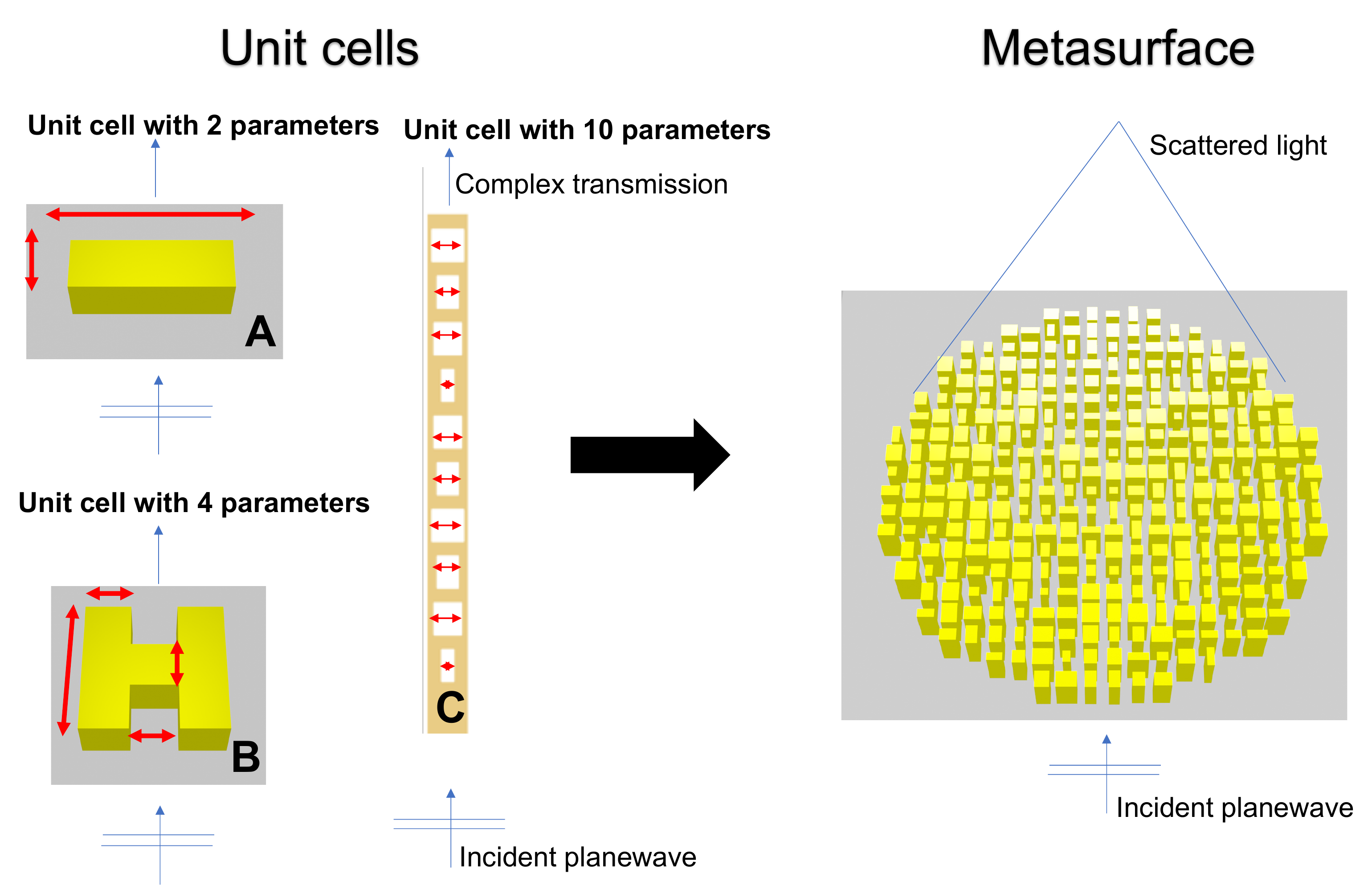}
\caption{(Left) Examples of three-dimensional and two-dimensional unit cells. 3D: fin unit cell with two parameters, H-shape unit cell with four parameters. 2D: multi-layer unit cell with holes with ten parameters. Each of the unit cell parameters are illustrated by red arrows. The transmitted field of the unit-cell is computed with periodic boundary conditions. When the period is subwavelength, the transmitted field can be summarized by a single complex number---the complex transmission. (Right) Unit cells (with independent sets of parameters) are juxtaposed to form a metasurface which is optimized to scatter light in a prescribed way. Using the local periodic approximation and the unit cell simulations, we can efficiently compute the approximate source equivalent to the metasurface and generate the field anywhere in the far-field.}
\label{fig:unitcellillustration}
\end{figure}

%%%Motivation
Designing metamaterials or composite materials, in which computational tools select composable components to recreate desired properties that are not present in the constituent materials, is a crucial task for a variety of areas of engineering (acoustic, mechanics, thermal/electronic transport, electromagnetism, and optics)~\cite{kadic20193d}.  For example in metalenses, the components are subwavelength scatterers on a surface, but the device diameter is often $>10^3$ wavelengths~\cite{khorasaninejad2017metalenses}.  Applications of such optical structures include ultra-compact sensors, imaging, and spectroscopy devices used in cell phone cameras and in medical applications~\cite{khorasaninejad2017metalenses}.  As the metamaterials become larger in scale and as the manufacturing capabilities improve, there is a pressing need for scalable computational design tools. 

In this work, surrogate models were used to rapidly evaluate the effect of each metamaterial components during device design~\cite{a15pestourie2018inverse}, and machine learning is an attractive technique for such models~\cite{an2019deep, jiang2019simulator, an2019generative, jiang2019free}.  However, in order to exploit improvements in nano-manufacturing capabilities, components have an increasing number of design parameters and training the surrogate models (using brute-force numerical simulations) becomes increasingly expensive.  The question then becomes: \emph{How can we obtain an accurate model from minimal training data?} We present a new \emph{active-learning}~(AL) approach---in which training points are selected based on an error measure (Fig.~\ref{fig:diagram})---that can reduce the number of training points by more than an order of magnitude for a neural-network (NN) surrogate model of partial-differential equations (PDEs). Further, we show how such a surrogate can be exploited to speed up large-scale engineering optimization by $> 100\times$. In particular, we apply our approach to the design of optical metasurfaces: large ($10^2$--$10^6$ wavelengths~$\lambda$) aperiodic nanopattered ($\ll \lambda)$ structures that perform functions such as compact lensing~\cite{1a3yu2011light}.  

Metasurface design can be performed by breaking the surface into unit cells with a few parameters each (Fig.~\ref{fig:unitcellillustration}) via domain-decomposition approximations~\cite{a15pestourie2018inverse, lin2019overlapping}, learning a ``surrogate'' model that predicts the transmitted optical field through each unit as a function of an \emph{individual} cell's parameters, and optimizing the total field (e.g. the focal intensity) as a function of the parameters of \emph{every} unit cell~\cite{a15pestourie2018inverse} (\secref{surrogate}).  This makes metasurfaces an attractive application for machine learning (\secref{algorithm}) because the surrogate unit-cell model is re-used millions of times during the design process, amortizing the cost of training the model based on expensive ``exact'' Maxwell solves sampling many unit-cell parameters.  For modeling the effect of a $1$--$4$ unit-cell parameters, Chebyshev polynomial interpolation can be very effective~\cite{a15pestourie2018inverse}, but encounters an exponential ``curse of dimensionality'' with more parameters~\cite{boyd2001chebyshev, trefethen2019approximation}.  In this paper, we find that a NN can be trained with orders of magnitude fewer Maxwell solves for the same accuracy with $\sim 10$ parameters, even for the most challenging case of multi-layer unit cells many wavelengths ($> 10\lambda$) thick (\secref{results}). In contrast, we show that subwavelength-diameter design regions (considered by several other authors~\cite{an2019deep, jiang2020deep, jiang2019simulator, ma2018deep, an2019generative, jiang2019free}) require orders of magnitude fewer training points for the same number of parameters (\secref{baseline}), corresponding to the physical intuition that wave propagation through subwavelength regions is effectively determined by a few ``homogenized'' parameters~\cite{holloway2011characterizing}, making the problems effectively low-dimensional. In contrast to typical machine-learning applications, constructing surrogate models for physical model such as Maxwell's equations corresponds to interpolating \emph{smooth} functions with no noise, and this requires new approaches to training and active learning as described in \secref{algorithm}.  We believe that these methods greatly extend the reach of surrogate model for metamaterial optimization and other applications requiring moderate-accuracy high-dimensional smooth interpolation.

% Previous metasurface design, previous applications of neural nets in optics [Raphael].
Recent work has demonstrated a wide variety of optical-metasurface design problems and algorithms. Different applications~\cite{3maguid2016photonic} such as holograms~\cite{5sung2019single}, polarization-~\cite{ 10arbabi2015dielectric, 11mueller2017metasurface}, wavelength-~\cite{14ye2016spin}, depth-of-field-\cite{bayati2020inverse}, or incident angle-dependent functionality~\cite{18liu2018metasurface} are useful for imaging or spectroscopy~\cite{23a6aieta2015multiwavelength,24zhou2018multilayer}.  \citeasnoun{a15pestourie2018inverse} introduced an optimization approach to metasurface design using Chebyshev-polynomial surrogate model, which was subsequently extended to topology optimization ($\sim 10^3$ parameters per cell) with ``online'' Maxwell solvers~\cite{lin2019topology}. Metasurface modeling can also be composed with signal/image-processing stages for optimized ``end-to-end design''~\cite{sitzmann2018end, lin2020end}. Previous work demonstrated NN surrogate models in optics for a few parameters~\cite{liu2018training, malkiel2018plasmonic, peurifoy2018nanophotonic}, or with more parameters in deeply subwavelength design regions~\cite{an2019deep, jiang2020deep}. As we will show in~\secref{baseline}, deeply subwavelength regions pose a vastly easier problem for NN training than parameters spread over larger diameters. Another approach involves generative design, again typically for subwavelength~\cite{an2019generative, jiang2019free} or wavelength-scale unit cells~\cite{liu2020topological}, in some cases in conjunction with larger-scale models~\cite{jiang2019simulator, ma2018deep, jiang2020deep}. A generative model is essentially the inverse of a surrogate function: instead of going from geometric parameters to performance, it takes the desired performance as an input and produces the geometric structure, but the mathematical challenge appears to be closely related to that of surrogates.

Active learning (AL) is connected with the field of uncertainty quantification (UQ), because AL consists of adding the ``most uncertain'' points to training set in an iterative way (\secref{algorithm}) and hence it requires a measure of uncertainty.   Our approach to UQ (\secref{algorithm}) is based on the NN-ensemble idea of \citeasnoun{lakshminarayanan2017simple} due to its scalability and reliability. There are many other approaches for UQ~\cite{hullermeier2019aleatoric,gal2016dropout,settles2009active,thiagarajan2020calibrating, chen2019confidence}, but \citeasnoun{lakshminarayanan2017simple} demonstrated performance and scalability advantages of the NN-ensemble approach.
In contrast, Bayesian optimization relies on Gaussian processes that scale poorly ($\sim N^3$ where $N$ is the number of training samples)~\cite{lookman2019active, bassman2018active}. To our knowledge, the work presented here is the first to achieve training time efficiency (we show an order of magnitude reduction sample complexity), design time efficiency (the actively learned surrogate model is at least two orders of magnitude faster than solving Maxwell’s equations), and realistic large-scale designs (due to our optimization framework~\cite{a15pestourie2018inverse}), all in one package.

%paragraph with our approach and summary results 

\section{Metasurfaces and surrogate models}\label{sec:surrogate}

% Explain figure 1 in more detail.  LPA.  $t(p) \approx \tilde{t}(p)$. 
In this section, we present the neural-network surrogate model used in this paper, for which we adopt the metasurface design formulation from \citeasnoun{a15pestourie2018inverse}.  The first step of this approach is to divide the metasurface into unit cells with a few geometric parameters $p$ each.   For example, Fig.~\ref{fig:unitcellillustration}(left) shows several possible unit cells: (a) a rectangular pillar (``fin'') etched into a 3d dielectric slab~\cite{a12khorasaninejad2016metalenses} (two parameters); (b) an H-shaped hole (four parameters) in a dielectric slab~\cite{an2019deep}; or a (c) multi-layered 2d unit cell with ten holes of varyings widths considered in this paper.   As depicted in Fig.~\ref{fig:unitcellillustration}(right), a metasurface consists of an array of these unit cells.  The second step is to solve for the transmitted field (from an incident planewave) \emph{independently} for each unit cell using approximate boundary conditions~\cite{a15pestourie2018inverse, lin2019topology, a12khorasaninejad2016metalenses, a4yu2014flat}, in our case a locally periodic approximation (LPA) based on the observation that optimal structures often have parameters that mostly vary slowly from one unit cell to the next~\cite{a15pestourie2018inverse}.  (Other approximate boundary conditions are also possible~\cite{lin2019overlapping}.) For a subwavelength period, the LPA transmitted far field is entirely described by a single number---the complex transmission coefficient $t(p)$.   One can then compute the field anywhere above the metasurface by convolving these approximate transmitted fields with a known Green's function, a near-to-farfield transformation~\cite{harrington}.   Finally, any desired function of the transmitted field, such as the focal-point intensity, can be optimized as a function of the geometric parameters of each unit cell~\cite{a15pestourie2018inverse}.

In this way, optimizing an optical metasurface is built on top of evaluating the function $t(p)$ (transmission through a \emph{single} unit cell as a function of its geometric parameters) thousands or even millions of times---once for every unit cell, for every step of the optimization process. Although it is possible to solve Maxwell's equations ``online'' during the optimization process, allowing one to use thousands of parameters $p$ per unit cell requires substantial parallel computing clusters~\cite{lin2019topology}.  Alternatively, one can solve Maxwell's equations ``offline'' (before metasurface optimization) in order to \emph{fit} $t(p)$ to a \emph{surrogate} model
\begin{equation}
    \tilde{t}(p) \approx t(p) \, ,
\end{equation}
which can subsequently be evaluated rapidly during metasurface optimization (perhaps for many different devices).  For similar reasons, surrogate (or ``reduced-order'') models are attractive for any design problem involving a composite of many components that can be modeled separately~\cite{an2019generative, jiang2019free, mignolet2013review}.  The key challenge of the surrogate approach is to increase the number of design parameters, especially in non-subwavelength regions as discussed in \secref{baseline}.

% Explain the neural net surrogate model in this paper (but not the error measure or active learning): only output so far is complex transmission.  Mention more outputs later.
% --> it is difficult to not mention the variance, because the baseline comparison because the objective function optimizes both the mean and the variance, even though we only use the mean in the baseline case.
In this paper, the surrogate model for each of the real and imaginary parts of the complex transmission is an ensemble of $J=5$ independent neural networks (NNs) with the same training data but different random ``batches''~\cite{goodfellow2016deep} on each training step. Each of NN~$i$ is trained to output a prediction $\mu_i(p)$ and an error estimate $\sigma_i(p)$ for every set of parameters $p$. To obtain these $\mu_i$ and $\sigma_i$ from training data $y(p)$ (from brute-force ``offline'' Maxwell solves) we minimize~\cite{lakshminarayanan2017simple}:
\begin{equation}\label{eq:loglikelihood}
    -\sum_p \log{p_{\Theta_i}(y|p)} = \sum_p \left[ \log{\sigma_i(p)} + \frac{(y(p)-\mu_i(p))^2}{2 \sigma_i(p)^2} \right]
\end{equation}
over the parameters $\Theta_i$ of NN~$i$.   \Eqref{loglikelihood} is motivated by problems in which $y$ was sampled from a Gaussian distribution for each $p$, in which case $\mu_i$ and $\sigma_i^2$ could be interpreted as mean and hetero-skedastic variance, respectively~\cite{lakshminarayanan2017simple}.  Although our exact function $t(p)$ is smooth and noise-free, we find that \eqref{loglikelihood} still works well to estimate the fitting error, as demonstrated in \secref{algorithm}.  Each NN is composed of an input layer with 13~nodes (10~nodes for the geometry parameterization and 3~nodes for the one-hot encoding~\cite{goodfellow2016deep} of three frequencies of interest), three fully-connected hidden layers with 256 rectified linear units (ReLU~\cite{goodfellow2016deep}), and one last layer containing one unit with a scaled hyperbolic-tangent activation function~\cite{goodfellow2016deep} (for~$\mu_i$) and one unit with a softplus activation function~\cite{goodfellow2016deep} (for~$\sigma_i$). Given this ensemble of $J$~NNs, the final prediction $\mu_*$ (for the real or imaginary part of $t(p)$) and its associated error estimate $\sigma_*$ are amalgamated as~\cite{lakshminarayanan2017simple}:
\begin{eqnarray}\label{eq:pooledm}
    \mu_*(p) = \frac{1}{\mathrm{J}}\sum_{i=1}^\mathrm{J} \mu_i(p)\\
    \sigma_*^2(p) = \frac{1}{\mathrm{J}}\sum_{i=1}^\mathrm{J} (\sigma_i^2(p) + \mu_i^2(p)) - \mu_*^2(p) \label{eq:pooledv}\, .
\end{eqnarray}

\section{Subwavelength is easier: Effect of diameter}\label{sec:baseline}

\begin{figure}[h!]
\centering
\includegraphics[width=\textwidth]{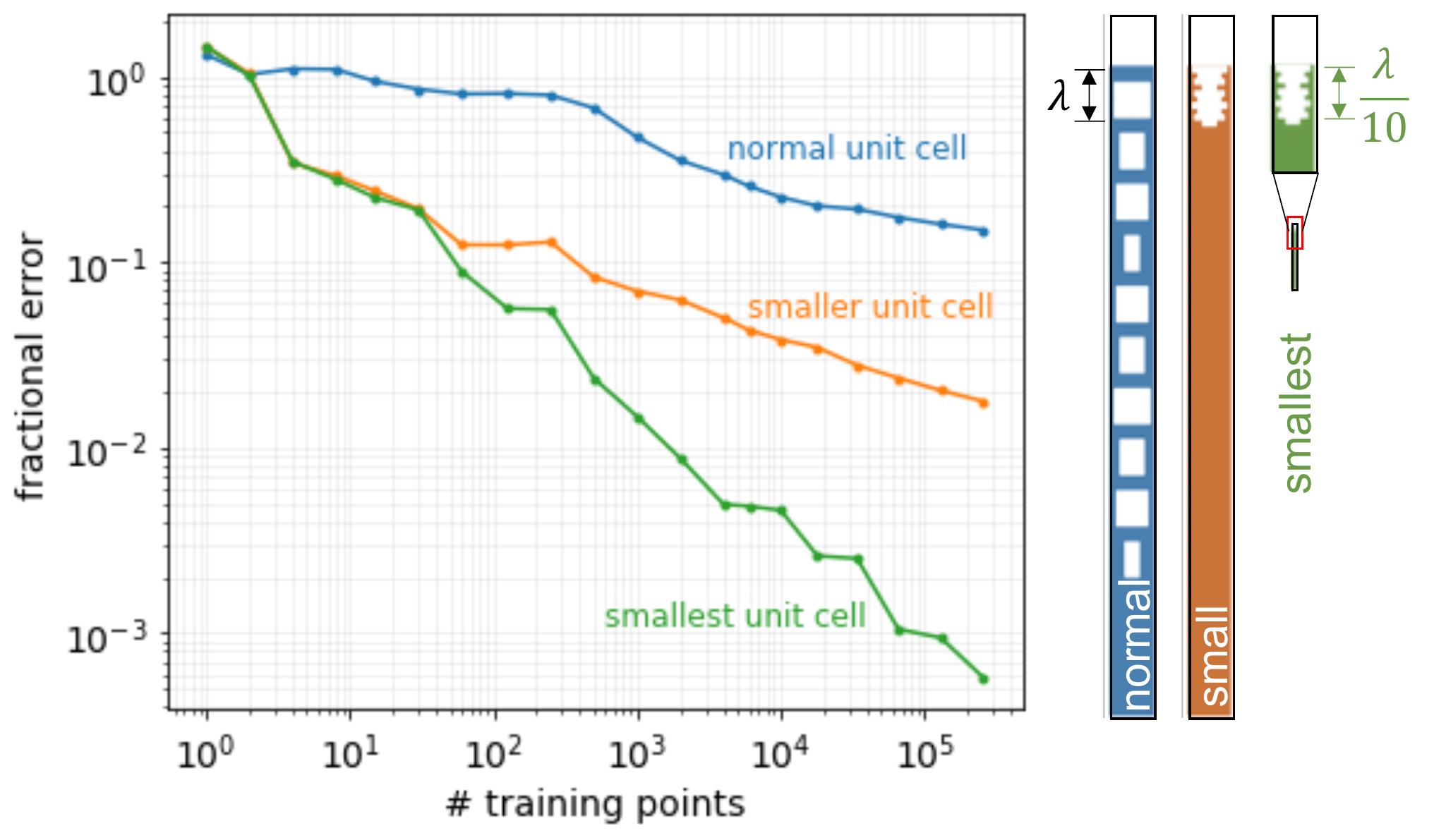}
\caption{Comparison of baseline training as we shrink the unit cell. Left: for the same number of training points, the fractional error (defined in Methods) on the test set of the small unit cell and the smallest unit cell are, respectively, one and two orders of magnitude better than the error of the main unit cell when using $1000$ training points or more, which indicates that parameters are more independent when the design-region diameter is big ($\gg \lambda$), and training the surrogate model becomes harder. Right: pictures of the unit cells to scale. Each color corresponds to the line color in the plot. For clarity, an inset shows the smallest unit cell enlarged 10 times.}
\label{fig:baselines}
\end{figure}

Before performing active learning, we first identify the regime where active learning can be most useful: unit-cell design volumes that are not small compared to the wavelength~$\lambda$. Previous work on surrogate models~\cite{an2019deep, jiang2020deep, jiang2019simulator, ma2018deep, an2019generative, jiang2019free} demonstrated NN surrogates (trained with random samples) for unit cells with $\sim 10^2$ parameters. However, these NN models were limited to a regime where the unit-cell degrees of freedom lay within a subwavelength-diameter volume of the unit cell.  To illustrate the effect of shrinking design volume on NN training, we trained our surrogate model for three unit cells (Fig.~\ref{fig:baselines}(right)): the main unit cell of this study is $12.5\lambda$ deep, the small unit cell is a vertically scaled-down version of the normal unit cell only $1.5\lambda$ deep, and the smallest unit cell is a version of the small unit cell further scaled down (both vertically and horizontally) by $10\times$.  Fig.~\ref{fig:baselines}(left) shows that, for the same number of training points, the fractional error (defined in Methods) on the test set of the small unit cell and the smallest unit cell are, respectively, one and two orders of magnitude better than the error of the main unit cell when using $1000$ training points or more. (The surrogate output is the complex transmission $\tilde{t}$ from \secref{surrogate}.)
That is, Fig.~\ref{fig:baselines}(left) shows that in the subwavelength-design regime, training the surrogate model is far easier than for larger design regions ($> \lambda)$.

Physically, for extremely sub-wavelength volumes the waves only ``see'' an averaged \emph{effective} medium~\cite{holloway2011characterizing}, so there are effectively only a few independent design parameters regardless of the number of geometric degrees of freedom.  Quantitatively, we find that the Hessian of the trained surrogate model (second-derivative matrix) in the smallest unit-cell case is dominated by only two singular values---consistent with a function that effectively has only two free parameters---with the other singular values being more than $100\times$ smaller in magnitude; for the other two cases, many more training points would be required to accurately resolve the smallest Hessian singular values.  A unit cell with large design-volume diameter ($\gg \lambda)$ is much harder to train, because the dimensionality of the design parameters is effectively much larger.

\section{Active-learning algorithm}\label{sec:algorithm}

\begin{figure}[h!]
\centering
\includegraphics[width=\textwidth]{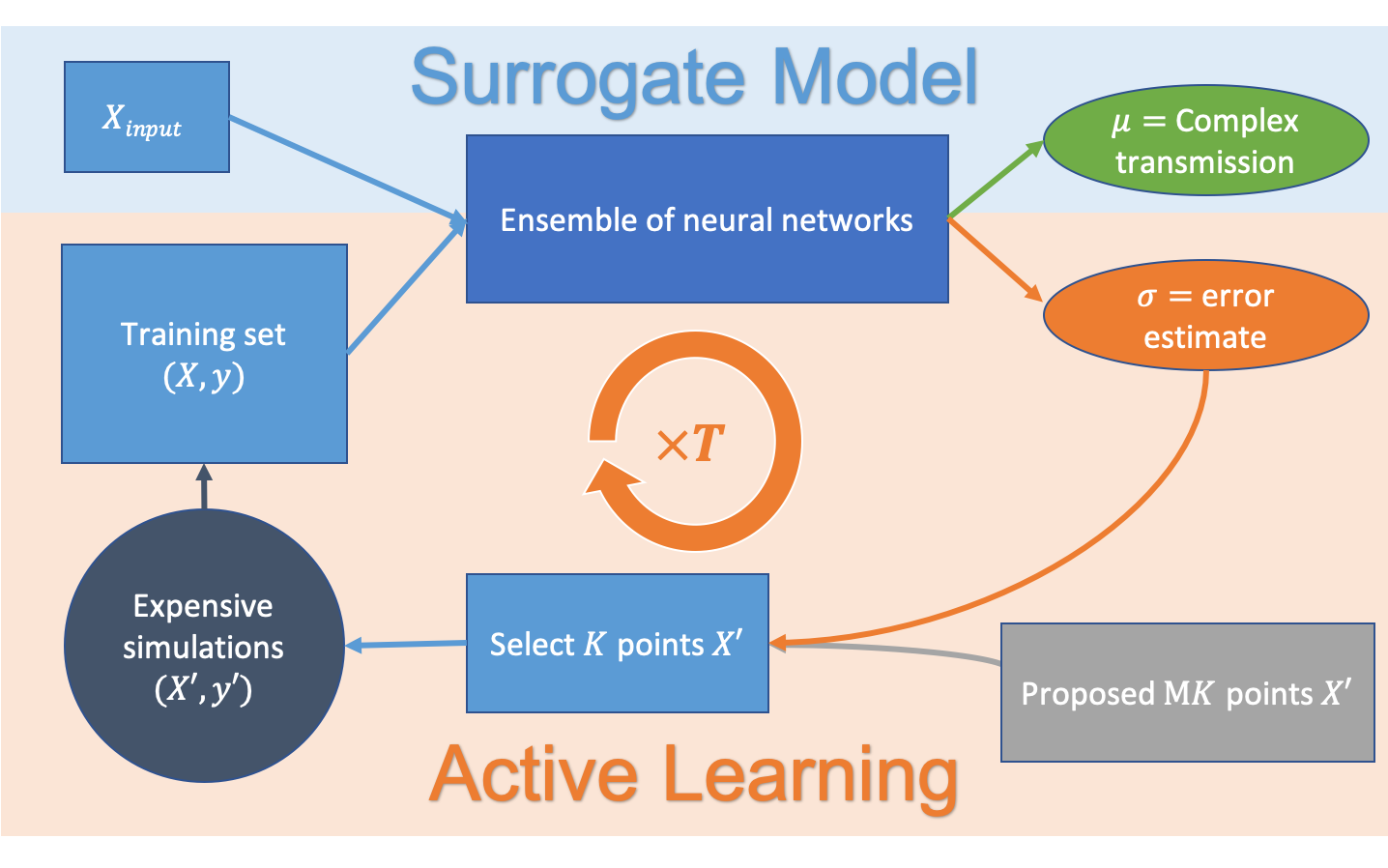}
\caption{Diagram of the surrogate model (blue background), and the  active-learning algorithm (orange background), the circle arrow signifies that the algorithm iterates T times. The fast evaluation of the surrogate is used both to create predictions of the surrogate model, and to compute the error measure that selects the points to add to the training set.}
\label{fig:diagram}
\end{figure}

% explain error measure and Fig.~\ref{fig:diagram}
Here, we present an algorithm to choose training points that is significantly better at reducing the error than choosing points at random.  As described below, we select the training points where the \emph{estimated model error is largest}, given the estimated error $\sigma_*$ from \secref{surrogate}. 

The algorithm used to train each of the real and imaginary parts is outlined in Fig.~\ref{fig:diagram} and Algorithm~\ref{alg:al}.  Initially we choose $n_\mathrm{init}$ uniformly distributed random points $p_1, p_2, ..., p_{n_\mathrm{init}}$ to train a first iteration $\tilde{t}^0(p)$ over 50 epochs~\cite{goodfellow2016deep}. Then, given the model at iteration $i$, we evaluate $\tilde{t}^i(p)$ (which is orders of magnitude faster than the Maxwell solver) at $MK$ points sampled uniformly at random and choose the $K$ points that correspond to the largest $\sigma_*^2$.  We perform the expensive Maxwell solves only for these $K$ points, and add the newly labeled data to the training set.  We train $\tilde{t}^{i+1}(p)$ with the newly augmented training set. We repeat this process $T$ times.

Essentially, the method works because the error estimate $\sigma_*$ is updated every time the model is retrained with an augmented dataset.  In this way, model tells us where it does poorly by setting a large $\sigma_*$ for parameters $p$ where the estimation would be bad in order to minimize \eqref{loglikelihood}.   %We discuss possible improvements to the error estimate in \secref{conclusion}, but our results in \secref{results} already demonstrate an order-of-magnitude reduction in training time compared to uniform random sampling (the ``baseline'' algorithm).

\begin{algorithm}[H]\label{alg:al}
\SetAlgoLined
\KwResult{the surrogate model $\tilde{t}(p)$ ($\mu_*$ and $\sigma_*$)}
 $\mathrm{P_0}$ = $n_\mathrm{init}$ points chosen at random \;
 Solve expensive PDE for each points in $\mathrm{P_0}$\;
 Create the first iteration of the labeled training set $\mathrm{TS}_0$\;
 Train the ensemble $\tilde{t}^0(p)$ on  $\mathrm{TS}_0$\;
 \For{i = 1:$T$}{
 $\mathrm{R_i}$ = M$\times$K points chosen at random \;
 Compute (cheaply) the error measures $\sigma^{i-1}_*(p)$ using $\tilde{t}^{i-1}$, $\forall~p \in \mathrm{R_i}$\;
 $\mathrm{P_i}$ = select K points in $R_i$ with the highest error measures  $\sigma^{i-1}_*$\;
 Solve expensive PDE for each points in $\mathrm{P_i}$ and get $t(p)$, $\forall~p \in \mathrm{P_i}$\;
 Augment the labeled training set with new labeled data $\mathrm{TS}_i$\;
 Train the ensemble $\tilde{t}^i(p)$ on  $\mathrm{TS}_i$\;
 }
 \caption{Active-learning of the surrogate model}
\end{algorithm}

\section{Active-learning results}\label{sec:results}

\subsection{Order-of-magnitude reduction in training data}

\begin{figure}[h!]
\centering
\includegraphics[width=.7\textwidth]{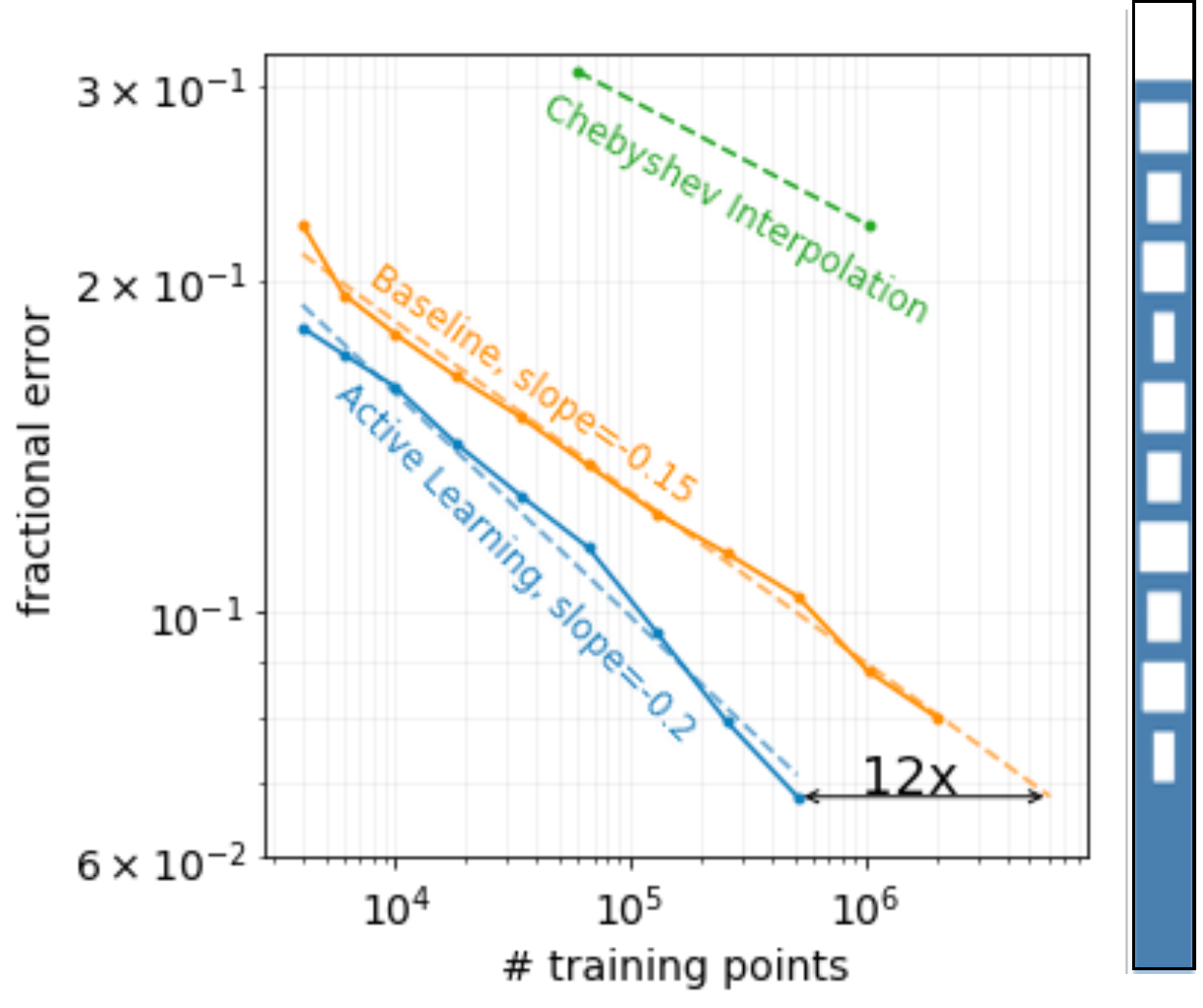}
\caption{(Left) The lower the desired fractional error, the greater the reduction in training cost compared to the baseline algorithm; the slope of the  active-learning fractional error ($-0.2$) is about 30\% steeper that that of baseline ($-0.15$).  The  active-learning algorithm achieves a reasonable fractional error of $0.07$ in twelve times less points than the baseline, which corresponds to more than one order of magnitude saving in training data. Chebyshev interpolation (surrogate for blue frequency  only) does not compete well with this number of training points. (Right) Unit cell corresponding to the surrogate model.}
\label{fig:al_vs_baseline}
\end{figure}

We compared the fractional errors of a NN surrogate model trained using uniform random samples with an identical NN trained using an active-learning approach, in both cases modeling the complex transmission of a multi-layer unit cell with ten independent parameters (Fig.~\ref{fig:al_vs_baseline}(right)).  With the notation of \secref{algorithm}, the baseline corresponds to $T=0$, and $n_{\mathrm{init}}$ equal to the total number of training points.  This corresponds to no active learning at all, because the $n_{\mathrm{init}}$ points are chosen at random. In the case of active learning, $n_\mathrm{init}=2000$, $M=4$, and we computed for $K=500,$ $1000,$ $2000,$ $4000,$ $8000,$ $16000,$ $32000,$ $64000,$ and $128000$.  Although three orders of magnitude on the log-log plot is too small to determine if the apparent linearity indicates a power law, Fig.~\ref{fig:al_vs_baseline}(left) shows that \emph{the lower the desired fractional error, the greater the reduction in training cost} compared to the baseline algorithm; the slope of the  active-learning fractional error ($-0.2$) is about 30\% steeper that that of baseline ($-0.15$).  The  active-learning algorithm achieves a reasonable fractional error of $0.07$ in twelve times less points than the baseline, which corresponds to more than one order of magnitude saving in training data (much less expensive Maxwell solves). This advantage would presumably increase for a lower error tolerance, though computational costs prohibited us from collecting orders of magnitude more training data to explore this in detail. For comparison and completeness, Fig.~\ref{fig:al_vs_baseline}(left) shows fractional errors using Chebyshev interpolation (for the blue frequency only). Chebyshev interpolation has a much worse fractional error for a similar number of training points. Chebyshev interpolation suffers from the ``curse of dimensionality''---the number of training points is exponential with the number of variables. The two fractional errors shown are for three and four interpolation points in each of the dimensions, respectively. In contrast, NNs are known to mitigate the ``curse of dimensionality''~\cite{cheridito2019efficient}.

\begin{figure}[ht!]
\centering
\includegraphics[width=0.75\textwidth]{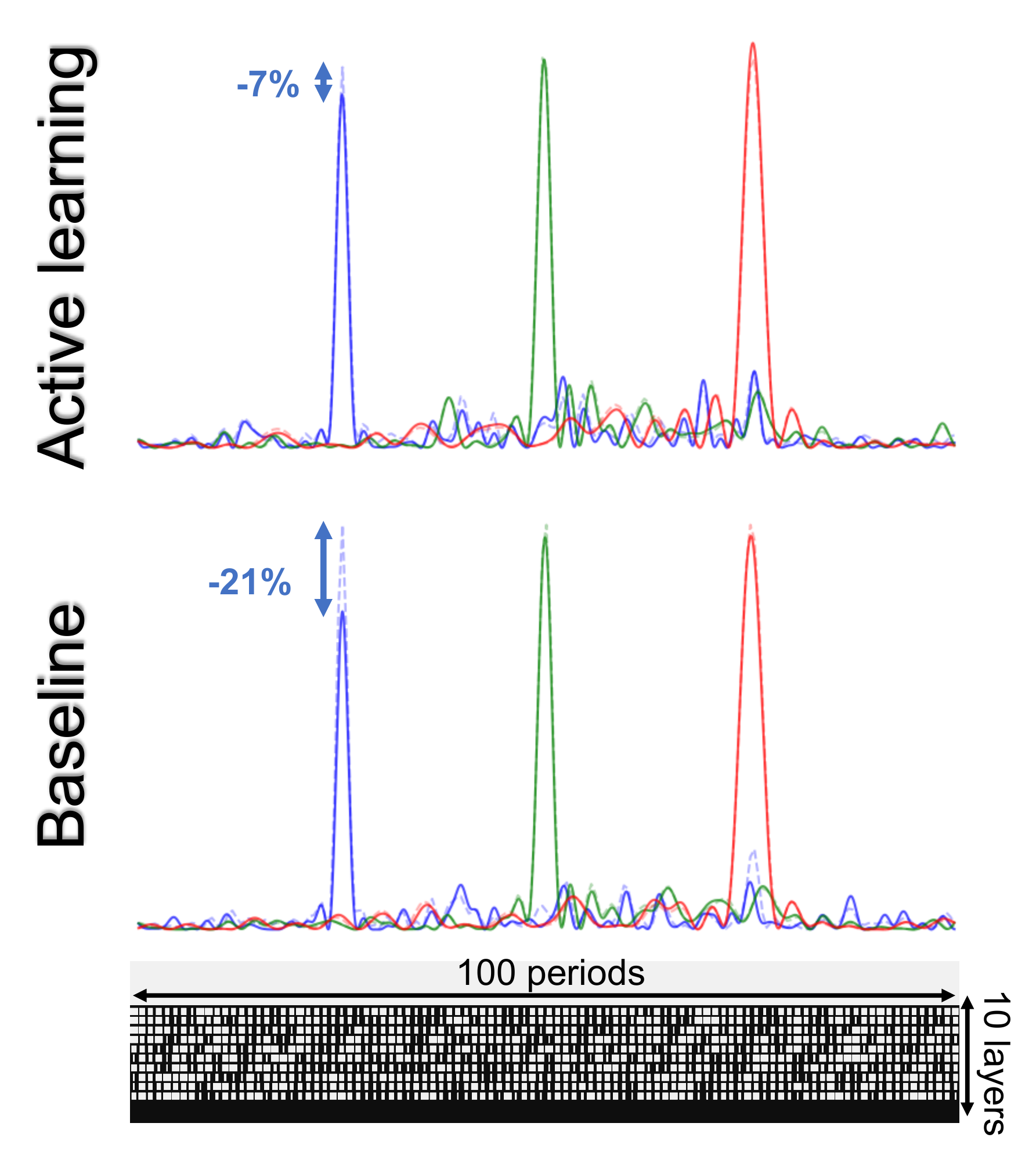}\caption{(Top) We used the active-learning and the baseline surrogates models to design a multiplexer---an optical device that focuses different wavelength at different points in space.  The actively learned surrogate model results in a design that much more closely matches a numerical validation than the baseline surrogate. This shows that the  active-learning surrogate is better at driving the optimization away from regions of inaccuracy. (Bottom) The resulting metastructure for the  active-learning surrogate with 100 unit cells of 10 independent parameters each (one parameter per layer).}
\label{fig:designs}
\end{figure}

\subsection{Application to metalens design}
We used both surrogates models to design a multiplexer---an optical device that focuses different wavelength at different points in space.  The actively learned surrogate model results in a design that much more closely matches a numerical validation than the baseline surrogate (Fig.~\ref{fig:designs}).  As explained in \secref{surrogate}, we replace a Maxwell's equations solver with a surrogate model to rapidly compute the optical transmission through each unit cell; a similar surrogate approached could be used for optimizing many other complex physical systems. In the case of our two-dimensional unit cell, the surrogate model is two orders of magnitude faster than solving Maxwell's equations with a finite difference frequency domain (FDFD) solver~\cite{champagne2001fdfd}.  The speed advantage of a surrogate model becomes drastically greater in three dimensions, where PDE solvers are much more costly while a surrogate model remains the same.

The surrogate model is evaluated millions of times during a meta-structure optimization. We used the actively learned surrogate model and the baseline surrogate model (random training samples), in both cases with $514000$ training points, and we optimized a ten-layer metastructure with $100$ unit cells of period $400$~nm for a multiplexer application---where three wavelengths (blue: $405$~nm, green: $540$~nm, and red: $810$~nm) are focused on three different focal spots ($-10~\mu$m, $60~\mu$m), ($0$, $60~\mu$m), and ($+10~\mu$m, $60~\mu$m), respectively.  The diameter is $40~\mu$m and the focal length is $60~\mu$m, which corresponds to a numerical aperture of $0.3$.  Our optimization scheme tends to yield results robust to manufacturing errors~\cite{a15pestourie2018inverse} for two reasons:  first, we optimize for the worst case of the three focal spot intensities, using an epigraph formulation~\cite{a15pestourie2018inverse};  second, we compute the average intensity from an ensemble of surrogate models that can be thought of as a Gaussian distribution $\tilde{t}(p) = \mu_*(p) + \sigma_*(p) \epsilon$ with $\epsilon \sim \mathcal{N}(0,1)$, and $\mu_*$ and $\sigma_*$ are defined in \eqref{pooledm} and \eqref{pooledv}, respectively,
\begin{equation}
    \mathbb{E} \left|E(\mathbf{r})\right|^2 = \left|\int G\mu_*\right|^2 + \left|\int G\sigma_*\right|^2
\end{equation} where $G$ is a Green's function that generates the far-field from the sources of the metastructure~\cite{a15pestourie2018inverse}.
% starting from Eq.~\ref{eq:green} and Eq.~\ref{eq:gaussian}, 
The resulting optimized structure for the  active-learning surrogate is shown in Fig.~\ref{fig:designs}(bottom).

In order to compare the surrogate models, we validate the designs by computing the optimal unit cell fields directly using a Maxwell solver instead of using the surrogate model. This is computationally easy because it only needs to be done once for each of the $100$ unit cells instead of millions of times during the optimization.  The focal lines---the field intensity along a line parallel to the two-dimensional metastructure and passing through the focal spots---resulting from the validation are exact solutions to Maxwell's equations assuming the locally periodic approximation (\secref{surrogate}).  Fig.~\ref{fig:designs}(top) shows the resulting focal lines for the  active-learning and baseline surrogate models.  A multiplexer application requires similar peak intensity for each of the focal spots, which is achieved using worst case optimization~\cite{a15pestourie2018inverse}.  Fig.~\ref{fig:designs}(top) shows that the actively learned surrogate has $\approx 3\times$ smaller error in the focal intensity compared to the baseline surrogate model.  This result shows that not only is the  active-learning surrogate more accurate than the baseline surrogate for $514000$ training points, but also the results are more robust using the  active-learning surrogate---the optimization does not drive the parameters towards regions of high inaccuracy of the surrogate model.  Note that we limited the design to a small overall diameter ($100$ unit cells) mainly to ease visualization (Fig.~\ref{fig:designs}(bottom)), and we find that this design can already yield good focusing performance despite the small diameter. In earlier work, we have already demonstrated that our optimization framework is scalable to designs that are orders of magnitudes larger~\cite{pestourie2020assume}.

%Previous work: combining active learning with the design \cite{chen2020generative} 
Previous work, such as   \citeasnoun{chen2020generative}---in a different approach to active-learning that does not quantify uncertainty---suggested iteratively adding the optimum design points to the training set (re-optimizing before each new set of training points is added).  However, we did not find this approach to be beneficial in our case.  In particular, we tried adding the data generated from LPA validations of the optimal design parameters, in addition to the points selected by our active learning algorithm, at each training iteration, but we found that this actually destabilized the learning and resulted in designs qualitatively worse than the baseline.  
%The validation points might bias the model to fit better in clusters where the model is already performing well, which introduces overfitting. 
By exploiting validation points, it seems that the active learning of the surrogate tends to explore less of the landscape of the complex transmission function, and hence leads to poorer designs. Such exploitation--exploration trade-offs are known in the active-learning literature~\cite{hullermeier2019aleatoric}.

\section{Concluding remarks}
\label{sec:conclusion}

In this paper, we present an  active-learning algorithm for composite materials which reduces the training time of the surrogate model for a physical response, by at least one order of magnitude. The simulation time is reduced by at least two orders of magnitude using the surrogate model compared to solving the partial differential equations numerically. While the domain-decomposition method  used here is the locally periodic approximation and the partial differential equations are the Maxwell equations, the proposed approach is directly applicable to other domain-decomposition methods (e.g.~overlapping domain approximation~\cite{lin2019overlapping}) and other partial differential equations or ordinary differential equations~\cite{rackauckas2020universal}. 

We used an ensemble of NNs for interpolation in a regime that is seldom considered in the machine-learning literature---when the data is obtained from a smooth function rather than noisy measurements.  In this regime, it would be instructive to have a deeper understanding of the relationship between NNs and traditional approximation theory (e.g. with polynomials and rational functions~\cite{boyd2001chebyshev, trefethen2019approximation}). For example, the likelihood maximization of our method forces  $\sigma_*$ to go to zero when $\tilde{t}(p) = t(p)$.  Although this allows us to simultaneously obtain a prediction $\mu_*$ and an error estimate $\sigma_*$, there is a drawback.  In the interpolation regime (when the surrogate is fully determined), $\sigma_*$ would become identically zero even if the surrogate does not match the exact model away from the training points.  In contrast, interpolation methods such as Chebyshev polynomials yield a meaningful measure of the interpolation error even for exact interpolation of the training data~\cite{boyd2001chebyshev, trefethen2019approximation}. In the future, we plan to separate the estimation model and the model for the error measure using a meta-learner architecture~\cite{chen2019confidence}, with expectation that the meta-learner will produce a more accurate error measure and further improve training time. We believe that the method presented in this paper will greatly extend the reach of surrogate-model based optimization of composite materials and other applications requiring moderate-accuracy high-dimensional interpolation.
% Future work: make sure there is no overlap in information in the batch
% e.g. Boltzmann transport equation~\cite{nguyen2020nano}

\section*{Methods}
%as in https://www.nature.com/articles/s41524-020-00371-x
\subsection{Training-data computation}
The complex transmission coefficients were computed in parallel using an open-source finite difference frequency-domain solver for Helmholtz equation~\cite{pestourie2020fdfd} on a $3.5$~GHz $6$-Core Intel Xeon E$5$ processor. The material properties of the multi-layered unit cells are silica (refractive index of 1.45) in the substrate, and air (refractive index of 1) in the hole and in the background. In the normal unit cell, the period of the cell is 400~nm, the height of the ten holes is fixed to 304~nm and their widths varies between 60~nm and 340~nm, each hole is separated by 140~nm of substrate. In the small unit cell, the period of the cell is 400~nm, the height of the ten holes is 61~nm, and their widths varies between 60~nm and 340~nm, there is no separation between the holes. The smallest unit cell is the same as the small unit cell shrunk ten times (period of 40~nm, ten holes of heigth 6.1~nm and width varying between 6~nm and 34~nm).

\subsection{Metalens design problem}
The complex transmission data is used to compute the scattered field off a multi-layered metastructure with $100$ unit cells as in \citeasnoun{a15pestourie2018inverse}. The metastructure was designed to focus three wavelengths (blue: $405$~nm, green: $540$~nm, and red: $810$~nm) on three different focal spots ($-10~\mu$m, $60~\mu$m), ($0$, $60~\mu$m), and ($+10~\mu$m, $60~\mu$m), respectively. The epigraph formulation of the worst case optimization and the derivation of the adjoint method to get the gradient are detailed in \citeasnoun{a15pestourie2018inverse}. Any gradient based-optimization algorithm would work, but we used an algorithm based on conservative convex separable approximations~\cite{svanberg2002class}. The average intensity is derived from the distribution of the surrogate model $\tilde{t}(p) = \mu_*(p) + \sigma_*(p) \epsilon$ with $\epsilon \sim \mathcal{N}(0,1)$ and the computation of the intensity based on the local field as in \citeasnoun{a15pestourie2018inverse},
\begin{eqnarray*}
     |E(\mathbf{r})|^2 &=& \left|\int_\mathrm{\Sigma} G(\mathbf{r}, \mathbf{r}') (-\tilde{t}(p(\vec{r}')) \, d\mathbf{r}'\right|^2,\\
     &=& \int_\mathrm{\Sigma} \bar{G} (\bar{\mu}_*(p) + \bar{\sigma}_*(p) \epsilon) d\mathbf{r}' \int_\mathrm{\Sigma} G (\mu_*(p) + \sigma_*(p) \epsilon) d\mathbf{r}',\\
     &=& \int \bar{G} \bar{\mu}_* \int G\mu_* + \epsilon^2 \int \bar{G} \bar{\sigma}_* \int G\sigma_*+2\epsilon \mathrm{Re}\left(\int \bar{G} \bar{\mu}_*\int G\sigma_*\right),\\
     &=& \left|\int G\mu_*\right|^2 + \epsilon^2 \left|\int G\sigma_*\right|^2 + 2\epsilon \mathrm{Re}\left(\int \bar{G} \bar{\mu}_*\int G\sigma_*\right),
\end{eqnarray*} 
where the $\bar{(\cdot)}$ notation denotes the complex conjugate, the notations $\int_\mathrm{\Sigma} (\cdot) d\mathbf{r}'$ and $G(\mathbf{r}, \mathbf{r}')$ are simplified to $p$, $\int$ and $G$, and the notation $p(\vec{r}')$ is dropped for clarity.  From the linearity of expectation, 
\begin{eqnarray}
    \mathbb{E} \left|E(\mathbf{r})\right|^2 &=& \left|\int G\mu_*\right|^2 + \mathbb{E}(\epsilon^2) \left|\int G\sigma_*\right|^2 + 2\mathbb{E}(\epsilon) \mathrm{Re}\left(\int \bar{G} \bar{\mu}_*\int G\sigma_*\right),\\
    \mathbb{E} \left|E(\mathbf{r})\right|^2 &=& \left|\int G\mu_*\right|^2 + \left|\int G\sigma_*\right|^2,
\end{eqnarray} where we used that $\mathbb{E}(\epsilon)=0$ and $\mathbb{E}(\epsilon^2)=1$. 

\subsection{Active-learning architecture and training}
The ensemble of NN was implemented using PyTorch~\cite{paszke2017automatic} on a $3.5$~GHz $6$-Core Intel Xeon E$5$ processor. We trained an ensemble of $5$ NN for each surrogate models. Each NN is composed of an input layer with 13~nodes (10~nodes for the geometry parameterization and 3~nodes for the one-hot encoding~\cite{goodfellow2016deep} of three frequencies of interest), three fully-connected hidden layers with 256 rectified linear units (ReLU~\cite{goodfellow2016deep}), and one last layer containing one unit with a scaled hyperbolic-tangent activation function~\cite{goodfellow2016deep} (for~$\mu_i$) and one unit with a softplus activation function~\cite{goodfellow2016deep} (for~$\sigma_i$). The cost function is a Gaussian loglikelihood as in \eqref{loglikelihood}. The mean and the variance of the ensemble are the pooled mean and variance from \eqref{pooledm} and \eqref{pooledv}.  The optimizer is Adam~\cite{kingma2014adam}. The starting learning rate is $0.001$. After the tenth epoch, the learning rate is decayed by a factor of $0.99$. Each iteration of the active learning algorithm as well as the baseline were trained for $50$ epochs. The choice of training points is detailed in \secref{algorithm}.  The quantitative evaluations were computed using the fractional error on a test set containing $2000$ points chosen at random. The fractional error $FE$ between two vectors of complex values $\vec{u}_\mathrm{estimate}$ and $\vec{v}_\mathrm{true}$ is \begin{equation}
    FE = \frac{|\vec{u}_\mathrm{estimate}-\vec{v}_\mathrm{true}|}{|\vec{v}_\mathrm{true}|}
\end{equation} where $|\cdot|$ is the L$2$-norm for complex vectors.

\section*{Data Availability}

The data that support the findings of this study are available from the corresponding
author upon reasonable request.

\section*{References}
\bibliographystyle{naturemag}
\bibliography{references}

\section*{Acknowledgements}

This work was supported in part by  IBM Research, the MIT-IBM Watson AI Laboratory, the U.S. Army Research Office through the Institute for Soldier Nanotechnologies (under award~W911NF-13-D-0001), and by the PAPPA program of DARPA MTO (under award~HR0011-20-90016).

\section*{Author contributions}

 RP, YM, PD, and SGJ designed the study, contributed to the machine-learning approach, and analyzed results; RP led the code development, software implementation, and numerical experiments; RP and SGJ were responsible for the physical ideas and interpretation; TVN assisted in designing and implementing the training. All authors contributed to the algorithmic ideas and writing.

%on the idea to include gradient matching in the active learning algorithm. However this idea was not included in the paper, because it did not improve the performance, while requiring significantly more PDE solves (at best $2\times$) to evaluate the gradients of the training data.

\section*{Competing interests}
The authors declare no competing financial or non-financial interests.

\end{document}